\documentclass[conference]{IEEEtran}
\usepackage{graphicx}
\usepackage{cite}
\usepackage[utf8]{inputenc}
\usepackage{comment}
\usepackage{array}
\usepackage{longtable}
\usepackage{xspace} 
\usepackage{color}
\usepackage{float}
\usepackage{cancel}
\usepackage{multirow}
\usepackage{amsmath}
\usepackage{mathtools}
\usepackage{soul}
\usepackage{graphicx}
\usepackage{tabularx}
\usepackage{verbatim}
\usepackage{listings}
\usepackage{algorithm}
\usepackage{algorithmicx}
\usepackage{algpseudocode}
\IEEEoverridecommandlockouts
\usepackage{cite}
\usepackage{amsmath,amssymb,amsfonts}
\usepackage{graphicx}
\usepackage{textcomp}
\usepackage{xcolor}
\def\BibTeX{{\rm B\kern-.05em{\sc i\kern-.025em b}\kern-.08em
    T\kern-.1667em\lower.7ex\hbox{E}\kern-.125emX}}

\algnewcommand\algorithmicforeach{\textbf{for each}}
\algdef{S}[FOR]{ForEach}[1]{\algorithmicforeach\ #1\ \algorithmicdo}

\begin{document}

\title{Semantic Neighborhood Ordering in Multi-objective Genetic Programming based on Decomposition\\
}

\author{
\IEEEauthorblockN{Fergal Stapleton$^*$\thanks{$^*$Joint first authors}}
\IEEEauthorblockA{{Naturally Inspired Computation Research Group,} \\
{Department of Computer Science},
{Hamilton Institute}\\
Maynooth University, Ireland \\
fergal.stapleton.2020@mumail.ie}
\and
\IEEEauthorblockN{Edgar Galv\'an$^{*,+}$\thanks{$^+$Senior Author}}
\IEEEauthorblockA{{Naturally Inspired Computation Research Group,} \\
{Department of Computer Science},
{Hamilton Institute}\\
Maynooth University, Lero, Ireland \\
edgar.galvan@mu.ie}
}

\maketitle



\begin{abstract}
Semantic diversity in Genetic Programming has proved to be highly beneficial in evolutionary search. We have witnessed a surge in the number of scientific works in the area, starting first in discrete spaces and moving then to continuous spaces. The vast majority of these works, however, have focused their attention on single-objective genetic programming paradigms, with a few exceptions focusing on Evolutionary Multi-objective Optimization (EMO). The latter works have used well-known robust algorithms, including the Non-dominated Sorting
Genetic Algorithm II and the Strength Pareto Evolutionary Algorithm, both heavily influenced by the notion of Pareto dominance. These inspiring works led us to make a step forward in EMO by considering Multi-objective Evolutionary  Algorithms Based on Decomposition (MOEA/D). We show, for the first time, how we can promote semantic diversity in MOEA/D in Genetic Programming. 

\end{abstract}


\section{Introduction}
\label{sec:intro}


Genetic Programming (GP)~\cite{Koza:1992:GPP:138936} is a powerful Evolutionary Algorithm (EA)~\cite{EibenBook2003} that has been successfully applied in a variety of challenging problems including finding the automatic configuration of modern neural networks~\cite{galvan2021neuroevolution,galvan2021neuroevolutionb}, energy-based problems~\cite{galvan_neurocomputing_2015,Galvan_SAC_2014}.  Despite the popularity of GP and its proven effectiveness in the face of challenging problems, features such as deceptiveness and multiple local optima~\cite{Eiben:2015:nature}, it is also well-known that canonical GP has some limitations and researchers have developed new approaches to make it more reliable. To this end, researchers have focused their attention on neutrality~\cite{galvan2021neuroevolution,DBLP:conf/ppsn/LopezP06_2,DBLP:conf/eurogp/LopezDP08,10.1007/978-3-540-73482-6_9,DBLP:journals/tec/PoliL12,DBLP:journals/evs/LopezPKOB11}, locality~\cite{DBLP:conf/cec/LopezMOB10,Galvan-Lopez2011,Galvan-Lopez:2010:TUL:1830483.1830646}, reuse of code~\cite{DBLP:conf/eurogp/LopezPC04}, to mention some research topics of interest.

One of the most popular elements studied by GP researchers during recent years is semantics, with many works reporting substantial improvements of GP performance
achieved by encouraging semantic diversity during evolutionary runs. While multiple definitions of semantics have  been proposed in the GP community,
it is commonly accepted that semantics refers to the output of a GP program when it is executed on a (training) data set (also known as fitness cases).
We give a formal treatment of semantics in Section~\ref{sec:sub:semantics}.

Semantics can be classified into one of two main categories: indirect or direct. Indirect semantic approaches refer to those that act on the syntax (genotype) of GP individuals to indirectly increase semantic diversity. On the other hand, direct semantic approaches refer to those mechanisms that adapt genetic operators to work directly on the semantics of GP individuals. Both type of approaches have benefits and limitations. This work focuses on indirect semantic-based approaches.

To the best of our knowledge, however, there is no  study on the adoption of semantics in Evolutionary Multi-Objective Optimization (EMO) at large~\cite{Deb:2001:MOU:559152}, and in Multi-objective GP (MOGP), in particular, except from the first author's work~\cite{DBLP:conf/gecco/GalvanS19,Galvan-Lopez2016,Galvan_MICAI_2016, 9308386, stapleton2021msc, galvan2020promoting}. These six works have shed some light in EMO approaches that use the Pareto dominance concept. Concretely, Galv\'an et al. have used the Non-dominated Sorting Genetic Algorithm II (NSGA-II)~\cite{Deb02afast} and the Strength Pareto Evolutionary Algorithm (SPEA2)~\cite{Zitzler01spea2:improving}. A trial and error semantic-based crossover mechanism was also tested by the first authors~\cite{stapleton2021msc, galvan2020promoting} for Multi-objective Evolutionary Algorithm Based on Decomposition (MOEA/D)~\cite{4358754}, however this method does not naturally incorporate semantics into the MOEA/D framework. The main contribution of this work, is therefore to naturally incorporate semantics into MOEA/D methods in GP, specifically using the  Weighted Sum, Tchebycheff and Penalty-based Boundary Intersection  methods. Our proposed semantic-based method, dubbed semantic neighborhood ordering, yields better results on the average hypervolume of evolved Pareto-approximated fronts, Pareto optimal fronts as well as reducing dramatically the same number of solutions in the objective space.


The remainder of this paper is organized as follows. Section~\ref{sec:related} includes the related work. Section~\ref{sec:methodology} outlines the background in semantics and multi-objective genetic programming that allow us to present our present our approach. Section~\ref{sec:experimental} explains the experimental setup used and Section~\ref{sec:results} shows and discusses the results obtained by all the approaches adopted in this study. Finally, Section~\ref{sec:conclusions} concludes this paper.

\section{Related Work}
\label{sec:related}



Research into the use of semantics in genetic programming has seen a surge in uptake in recent years due to its noteworthy performance benefits. Such studies have looked at geometric operators~\cite{DBLP:conf/ppsn/MoraglioKJ12}, the analysis of indirect semantics~\cite{Galvan-Lopez2016,Uy2011} and the use of semantics in real-world problems~\cite{Vanneschi2013}. A summary of semantic works undertook by the GP community can be found in~\cite{Vanneschi:2014:SSM:2618052.2618082}. Broadly speaking semantics refers to the output of a GP program once it has been computed on a data set. One of the first major works to make use of semantics was the research by McPhee et al.~\cite{McPhee:2008:SBB:1792694.1792707} where the authors studied the semantics of subtrees. Moreover their study highlighted that the widely adopted 90-10 crossover operation produced semantically equivalent offspring for more than 75\% of crossover events that they tested. In other words crossover events do not contribute any immediate useful search in the semantic space of a problem.

Uy et al.~\cite{DBLP:conf/ae/UyOHML09} proposed an expensive Semantic Similarity-based Crossover  mechanism by determining if two expressions are semantically similar, with this method being tested on a series of symbolic regression problems. To determine if two expressions were semantically similar, a random set of points were sampled from the domain and used to calculate the sampling semantics. In turn the absolute differences of the sampling semantics for each expression is used to calculate semantic distance and if this value fell within predefined upper and lower bounds then the two expressions were deemed to be semantically similar and crossover was applied. A number of scenarios were tested where the expressions were either represented as two full program trees or two sub trees. One drawback of this method is that  in some cases these expressions require multiple trials before crossover was applied. 

To tackle this limitation, Galv\'an et al.~\cite{DBLP:conf/cec/LopezCTK13} proposed a more cost efficient mechanism based on the tournament selection operator. With this method the tournament selection of the first parent is selected as the fittest individual as chosen from a pool of individuals randomly selected from the population. The second parent is chosen from a pool of individuals that are semantically different from the first parent and additionally the second parent is also the fittest individual from its respective pool. If no individual is semantically different from the first parent, then the tournament selection of the second parent is performed as normal. The proposed approach alleviated the cost inefficient aspect of the original approach by Uy et al.~\cite{10.1007/978-3-642-04962-0_7, Uy2011, stapleton2021msc, galvan2020promoting} while producing comparable and in some cases better results.

A number of methods in recent times have made use of semantic distance~\cite{DBLP:conf/gecco/GalvanS19, 9308386} within the context of popular Evolutionary Multi-objective Objective (EMO) frameworks such as the Non-dominated Sorting Genetic Algorithm (NSGA-II)~\cite{Deb02afast} and Strength Pareto Evolutionary Algorithm (SPEA2)~\cite{Zitzler01spea2:improving}. These semantic-based methods\cite{DBLP:conf/gecco/GalvanS19, 9308386} make use of a reference individual referred to as a pivot to calculate the semantic distance between itself and every other individual in the population. The pivot is selected as an individual from the sparsest region of the first Pareto front. 
The semantic distance is used to replace the crowding distance and individuals which are semantically similar are retained. Furthermore, the semantic distance itself can be used as an additional criteria to optimize.

Recently, S. Ruberto et al.~\cite{10.1145/3377929.3397486} developed a semantic approach for symbolic regression called Semantic Genetic Programming Based on Dynamic Targets (SGP-DT). This method abandons classical crossover with the aim of minimizing bloat. At a top level overview the algorithm makes use of internal and external iterations. Internal iterations are synonymous to generational iterations in standard GP and return what is referred to as a partial model. The external population, on the other hand, modifies the training set and  replaces the estimated output with the residuals of the previous iterations partial model. This in turn results in the fitness function evaluating individuals differently for each partial model. The authors reasoned that as a result of the dynamic targets each partial model would focus on a different characteristic of the problem and that at each external iteration this characteristic influenced the fitness value the most.

The semantic based methods discuss so far focused on fitness selection methods that treated each objective separately. Unlike Pareto dominance-based methods which treat each objective separately when measuring fitness, Multi-objective Evolutionary Algorithm based on Decomposition (MOEA/D) decomposes the multi-objective problem into a subset of single objective scalar optimization problems~\cite{4358754}. Each single objective optimization problem in this context is defined using a weight vector. The weight vectors represent search directions in the objective space and are typically uniformly distributed to maintain diversity.


Decomposition methods have previously been applied to MOGP, but without considering semantics. S. Azari et al.~\cite{10.1007/978-3-030-41299-9_35} used a MOGP decomposition method for classification in tandem mass spectroscopy where the resulting spectra result in highly imbalanced data as the amount of noise (majority class) greatly outnumbers the number of signal peaks (minority class).

MOEA/D is not without its limitations. H. Sato~\cite{Sato2016chain} highlighted some of these key limitations. The author specified three issues inherent to the canonical algorithm

\begin{itemize}

\item In its canonical form duplication occurs when multiple neighbors are replaced by a single offspring with superior fitness. It was noted that for some problems this may improve convergence as offspring with better fitness are greedily selected for subsequent generations, however it may also lead to a loss in genetic diversity and hence reduce search performance.
\item Targets are predetermined before they are updated, that is, when an offspring is generated its parents are selected from neighboring individuals. It is, however, unknown if the original weights assigned to the targets are the optimal weights for the new offspring.
\item Existing solutions are updated on the fly, as soon as an offspring is determined to have better fitness. However, these removed solutions may still be beneficial for other weights and hence other search directions. 
\end{itemize}

\noindent The authors proposed methods addressing each of these limitations in turn. In part, we employ the same strategy for tackling duplication as outlined in the first point, but with an additional preference based mechanism that uses semantic ordering.

\section{Background and Methodology}
\label{sec:methodology}

\subsection{Semantics}
\label{sec:sub:semantics}

Pawlak et al.~\cite{6808504} gave a formal definition for program semantics. Let $p \in P$ be a program from a given programming language $P$. The program $p$ will produce a specific output $p(in)$ where input $in \in I$. The set of inputs $I$ can be understood as being mapped to the set of outputs $O$ which can be defined as $p:I \rightarrow O$. 

\textbf{ Def 1.} \emph{Semantic mapping function is a function $s:P \rightarrow S$ mapping any program $p$ from $P$ to its semantic $s(p)$, where we can show the semantic equivalence of two programs}. Eq~\ref{eq:def1} expresses this formally,

\begin{equation}
  s(p_1) = s(p_2) \iff  \forall\ in \in I: p_1(in) = p_2(in)
  \label{eq:def1}
\end{equation}

\noindent This definition presents three important and intuitive properties for semantics:

\begin{enumerate}
\item Every program has only one semantic attributed to it.
\item Two or more programs may have the same semantics.
\item Programs which produce different outputs have different semantics.
\end{enumerate}

\noindent In Def. 1, we have not given a formal representation of semantics. In the following, semantics will be represented as a vector of output values which are executed by the program under consideration using an input set of data. For this representation of semantics we need to define semantics under the assumption of a finite set of fitness cases, where a fitness case is a pair consisting of a program input and its respective program output $I$ $\times$ $O$. This allows us to define the semantics of a program as follows.

\textbf{Def 2.} \emph{The semantics $s(p)$ of a program $p$ is the vector of
values from the output set $O$ obtained by computing $p$ on all inputs from the input set $I$}. This is formally expressed in Eq.~\ref{eq:def2},

\begin{equation}
  s(p) = [p(in_1), p(in_2), ... , p (in_l)]
  \label{eq:def2}
\end{equation}

\noindent where $l = |I|$ is the size of the input set. \\

\subsection{MOEA/D Canonical}
\label{sec:sub:moead_canon}

Multi-objective Evolutionary Algorithm based on Decomposition (MOEA/D) decomposes a multi-objective problem into a subset of scalar optimization problems. 
The single objective optimization sub-problems are defined by the scalar optimization function $g$ using a uniform distribution of weight vectors $\lambda^i$ where $i = \{1, 2 ... N\}$. Weight assignment is based on the euclidean distance of the solutions after individuals have been initialized and each weight vector represents a direction in the search space. Each sub-problem is optimized based purely on information from its neighboring sub-problems and as such the exploration and exploitation aspects of the MOEA/D algorithm are dependent upon the neighborhood structure. In the original paper by Zhang et al.~\cite{4358754}, three scalarization decomposition methods were proposed: Weighted sum, Tchebycheff and Penalty-based Boundary Intersection (PBI). We briefly discuss each of these. \\

\noindent \textbf{Weighted Sum}\\

\noindent The scalar optimization of $g^{ws}$ is given by Eq.~\ref{eq:scalar}

\begin{equation}
\max(g^{ws}(x | \lambda)) = \sum_{i=1}^{m} \lambda_i f_i (x)
\label{eq:scalar}
\end{equation}

\begin{equation*}
\text{subject to } x \in \Omega
\end{equation*}

\noindent where $j =\{1, 2 ... m\}$ is the dimensions of the objective space, $f_i$ is the objective function, $\Omega$ is the decision (variable) space and $x$ is the variable to be optimized. The weighted sum approach has been noted as poorly approximating the optimal Pareto front with non-concave problems. \\

\noindent \textbf{Tchebycheff}\\

\noindent The scalar optimization of $g^{tch}$ is given by Eq.~\ref{eq:tche}

\begin{equation}
\min(g^{tch}(x|\lambda)) = \underset{1 \leq j \leq m}{\max}  \{\lambda | f_j(x) - z_{j} | \}
\label{eq:tche}
\end{equation}

\noindent The ideal point $z_j$ represents the best objective function value for $f_j$ found in the population thus far. \\

\noindent \textbf{Penalty-based Boundary Intersection (PBI)}\\

\noindent The scalar optimization of $g^{tch}$ is given by Eq.~\ref{eq:pbi}

\begin{equation}
\min(g^{pbi}(x|\lambda)) = d_1 + \theta d_2
\label{eq:pbi}
\end{equation}

\noindent where,

\begin{equation}
d_1 = \frac{|| (f(x)-z)^{T} \lambda ||}{|| \lambda ||} , \ 
d_2 = || (f(x) - (z - d_1 \frac{\lambda}{|| \lambda ||}) ||
\end{equation}

\noindent where $\theta$ is a predefined penalty parameter set by the user.

\noindent The steps involved in the algorithm are given below in more detail. \\

\noindent \textbf{Step 1:} Initialization.\\

\noindent \textbf{Step 1.1)} Initialize external population $EP = \emptyset$.\\

\noindent \textbf{Step 1.2)} Calculate the Euclidean distance between any two weight vectors and find the T closest weight vectors to each respective weight vector. For each $i = \{1, 2, ... ,N\}$, set $B(i) = \{i_1, i_2, ...  , i_T \}$ where $B(i)$ can be understood as a neighborhood reference table of indices and where $\lambda^{i_1} , \lambda^{i_2} , ..., \lambda^{i_T}$ are the $T$ closest weight vectors to $i$.\\

\noindent \textbf{Step 1.3)} Randomly create the initial population $x_1, x_2...x_N$ and calculate initial fitness value $F(x_i)$.\\

\noindent \textbf{Step 2:} Apply genetic operations and search for solutions.\\

\noindent \textbf{Step 2.1)} Select two indices $k$ and $l$ randomly from the neighborhood reference table $B(i)$ and generate new offspring $y$ from parents $x_k$ and $x_l$ by applying genetic
operations.\\

\noindent \textbf{Step 2.2)} Update $z$ such that for each $j = \{1, 2, ... ,m\}$ if $z_j < f_j(y)$, then set $z_j = f_j(y)$. In the case where the objective is to minimize $F(x)$, then this inequality is reversed.\\

\noindent \textbf{Step 2.3)} Update the neighboring solutions for the $j^{th}$ case such that $j \in B(i) $, if $g(y | \lambda^j, z) \geq g(x^j | \lambda^j, z)$, then let $x^j = y$ and calculate new fitness $F(y)$.\\

\noindent \textbf{Step 3}: Fill external population and termination.\\

\noindent \textbf{Step 3.1)} EP is filled with non-dominated solutions across all generations, where newly dominated solutions at each generation are removed. When stopping criteria is satisfied then the non-dominated solutions of EP are outputted; otherwise Step 2 is repeated.\\



Solution search as determined by each of the decomposition methods controls which population members are updated. In the standard update (Step 2.3), each program from the current population is updated with the child program. Since each index $j$ from the neighborhood reference table is checked multiple times against the child program this might lead to multiple replacements within the specified neighborhood. As such under the standard update, duplication of programs can occur at a generational level.

One way to alleviate this duplication is to break out of the loop after the first replacement has occurred. Such an approach would immediately reduce the amount of duplication occurring at each generation, though subsequently there is no guarantee that the most beneficial replacement would have occurred. To facilitate this pitfall, such an approach would need to order the neighborhood prior to undergoing the solution search.

\subsection{Semantics in MOEA/D}
\label{sec:sub:moead_semantics}

To this end, we propose using semantic distance to order the neighborhood first. Similar to how previous semantic distance-based approaches have calculated semantic distance this proposed method also makes use of a \textit{pivot}~\cite{DBLP:conf/gecco/GalvanS19, 9308386}. The pivot is selected as an individual from the sparsest region from the non-dominated solutions of the external population EP (Alg.~\ref{sem_alg}: line 7-8). The semantic distance is calculated as the absolute difference between the pivot $p$ as selected from the external population $EP$ and every individual $v$ from the standard population $P$ using Eq.~\ref{eq:dist}

\begin{equation}
d(p_j , v) = \sum_{i=1} 1\ if\ | p(in_{i}) - v(in_{i}) | < \text{UBSS}
\label{eq:dist}
\end{equation}

 As such, individuals that are semantically similar to the pivot will have a larger distance than those which are dissimilar. For the sake of simplicity, a single upper bound is chosen. The use of bounds in continuous spaces is common to quantify semantic diversity as profusely used in the specialized literature~\cite{DBLP:conf/ae/UyOHML09, DBLP:conf/gecco/GalvanS19, 9308386}.
 The neighborhood is sorted such that the most semantically dissimilar individuals are checked first (Alg.~\ref{sem_upd_alg}: line 2-4) and once a preferable update occurs the next child program is evaluated (Alg.~\ref{sem_upd_alg}).

\begin{algorithm}[t]
\caption{Semantically Ordered MOEA/D}\label{sem_alg}
\begin{algorithmic}[1]
\State $\Lambda = \{\lambda^{i_1}, \lambda^{i_2}, ..., \lambda^{i_N}\} \gets \text{Create weight vectors}$
\State $P = \{x_1, x_2, ..., x_N\} \gets \text{Initialize population}$
\ForEach {$ \lambda^{i} \in \Lambda$} 
\State $B(i) = \{i_1, i_2, ..., i_T\} \gets \text{Define reference table}$
\EndFor
\Repeat
\State $CD_1 \gets crowding\_distance(EP)$
\State $pivot \gets furthest\_point(CD_1)$
\ForEach {$ i \in \{1, 2, ..., N\} $} 
\State $k, l \gets \text{Return random parent indices from B(i)}$
\State $y \gets \text{Generate child program from } x^k \text{ and } x^l$
\State $P = Update(y, i, pivot)$
\EndFor
\State $EP\gets \text{Remove non-dominated solutions based on P}$ 
\Until{ Stopping criteria is met}
\State \Return {The non-dominated solutions from EP}
\end{algorithmic}
\end{algorithm}

\begin{algorithm}[t]
\caption{Semantically Ordered Update}\label{sem_upd_alg}
\begin{algorithmic}[1]
\ForEach {$ i \in T $}
\State $NP \gets \text{Subset population using B(i) and P}$
\State $NP \gets Compute\_Semantics(pivot, NP)$
\State $B(k) \gets Sort(B(i), NP)$
\EndFor
\ForEach {$ j \in B(k) $}
\If{$g(y | \lambda^j, z) > g(x^j | \lambda^j, z)$}
\State $x_j = y \gets \text{Single replacement}$
\State \Return {} $\gets \text{Exit subroutine}$
\EndIf
\EndFor
\end{algorithmic}
\end{algorithm}

\section{Experimental setup}
\label{sec:experimental}

The primary source for the data sets used during experimentation were from the well-known University of California, Irvine Machine Learning repository with data sets containing varying degrees of majority-to-minority class imbalance ratios and numbers of features~\cite{Dua:2019 ,gmd-6-1157-2013,Little_2007}, summary of which is given in Table \ref{tab:datasets}. These problems are of different nature and complexity. They have from a few features up to dozens of {them, including binary and real-valued features.} These benchmark problems are carefully selected to encompass a varied collection of problem domains to ensure that the evaluation of the approaches used in this work, including all the scalar optimization approaches used in this work; namely the Weighted Sum, Tchebycheff and Penalty-based Boundary Intersection methods, along with their corresponding semantic-based methods, are not problem or domain specific. The training and test data is split 50/50 for each data set maintaining the same class imbalance ratio for each split. All the results are reported on the test data.

\begin{table*}[t!]
\caption{Binary imbalanced classification data sets used in our research}
\centering
\resizebox{0.90\textwidth}{!}{ 
\begin{tabular}{llrrrrrr}
\hline
Data set & Classes             & \multicolumn{3}{c}{Number of examples} & Imb. & \multicolumn{2}{c}{Features} \\
 & Positive/Negative (Brief description) &  Total   & Positive    &  Negative     & Ratio  & No. & Type \\ \hline
Ion    & Good/bad (ionosphere radar signal)     & 351  & 126 (35.8\%) & 225 (64.2\%)       &1:3     &34  & Real  \\
Spect  & Abnormal/normal (cardiac tom. scan)   & 267  & 55 (20.6\%)  &212 (79.4\%)        &1:4     & 22  & Binary  \\
Yeast$_1$& mit/other (protein sequence)         &1482   & 244 (16.5\%) & 1238 (83.5\%)     &1:6     & 8 & Real  \\
Yeast$_2$& me3/other (protein sequence)         & 1482  & 163 (10.9\%) & 1319 (89.1\%)     &1:9     & 8 & Real  \\
Climate&  failure/success (simulation crashes)    & 540  & 46 (\ 8.5\%) & 494 (91.5\%)     & 1:12       &  18 & Real \\
Glass & building windows float proc./other (type of glass)     & 214   &  70 (32.7\%) & 144 (67.3\%)    &  1:3      &  10 & Real \\
Parkinson's & parkinson's disease/healthy (vocal measurements)     & 197  &  48 (24.6\%) & 147 (75.4\%)    &  1:4      &  22 & Real \\
Wine & wine type 1/other (Alcohol cultivar)     & 178   &  59 (33.1\%) & 119 (66.9\%)    &  1:3      &  12 & Real \\
\hline
\end{tabular}
}
\label{tab:datasets}
\end{table*}

The terminal and function sets used in this work are as follows. The terminals are the problem features. The function set is composed of arithmetic operators $ \Re = \{+, -, *, \%\} $, where \% denotes the protected division operator. These functions are used to build a classifier (mathematical expression) that returns a single value for a given input (data example to be classified). This number is mapped onto a set of class labels using zero as the class threshold. In our work, an example is assigned to the minority class if the output of the classifier is greater or equal to zero. It is assigned to the majority class, otherwise.

Classification accuracy is a common metric used in determining fitness for binary classification problems where $ACC = \frac{TP + TN}{TP + TN + FP + FN}$. However, with the imbalanced data sets such as the ones modeled in this work, using this accuracy measure will tend to bias towards the majority class as shown in~\cite{6198882}. A better approach is to treat the minority and majority as two separate objectives where the goal is to maximize the number of correctly classified cases. This can be done using the true positive rate $TPR = \frac{TP}{TP + TN}$ and true negative $TNR = \frac{TN}{TN + FP}$~\cite{6198882}. Table \ref{tab:parameters} gives an overview of the parameters used in our work. These values are the result of preliminary experiments not reported in this work. In particular, the value for the semantic threshold yields promising results. This is in agreement with other studies indicating that this value successfully promotes semantics helping evolutionary search~\cite{Galvan_MICAI_2016, DBLP:conf/gecco/GalvanS19, Uy2011}. Furthermore for the PBI method only a single penalty parameter value $\theta$ = 0.1 was employed.

To obtain meaningful statistical results, we executed 30 independent runs for each of the methods used in this work. This gives a total of 1,440 independent runs\footnote{30 independent runs, 8 data sets, 3 semantic-based MOEA/D approaches and 3 canonical MOEA/D methods (the Weighted Sum, Tchebycheff and Penalty-based Boundary Intersection).}.

\begin{table}
\centering
\caption{Summary of parameters}
\resizebox{0.90\columnwidth}{!}{ 
\small\begin{tabular}{|l|r|} \hline 
\emph{Parameter} &
\emph{Value} \\ \hline \hline
Population Size & 500 \\ \hline
Generations & 50 \\ \hline
Type of Crossover & 90\% internal nodes, 10\% leaves  \\ \hline
Crossover Rate  & 0.60  \\ \hline
Type of Mutation & Subtree \\ \hline
Mutation Rate & 0.40 \\ \hline
Selection & Tournament (size = 7) \\ \hline
Initialization Method & Ramped half-and-half \\ \hline
Initialization Depths: & \\ 
\hspace{.3cm}Initial Depth & 1 (Root = 0)\\ 
\hspace{.3cm}Final Depth & 5 \\ \hline
Maximum Length & 800 nodes \\ \hline
Maximum Final Depth & 8\\ \hline
Independent Runs & 30 \\ \hline
Semantic Thresholds &  UBSS = 0.5 \\ \hline
Neighborhood size & 20 \\ \hline
Scalar Optimization & \{ Weighted Sum, Tchebycheff \& PBI \} \\ \hline
PBI Theta Value & 0.1 \\ \hline
\end{tabular}
}
\label{tab:parameters}
\end{table}

\section{Results}
\label{sec:results}





\subsection{Hypervolume}
\label{sec:sub:hypervolume}

Tables \ref{tab:hypervolumn:moeadcannon} and \ref{tab:hypervolumn:moeadsemantics} reports the average hypervolume over 30 independent runs for the external population. We also computed the accumulated Pareto optimal (PO) front with respect to 30 runs from the external population, that is the set of non-dominated solutions after merging all 30 Pareto-approximated
fronts. These results were gathered for both the canonical MOEAD/D
framework (Table~\ref{tab:hypervolumn:moeadcannon}) and for the framework which incorporates the semantic neighborhood ordering (Table~\ref{tab:hypervolumn:moeadsemantics}), as outlined in section \ref{sec:sub:moead_semantics}. The Weighted Sum (WGT), Tchebycheff (TCH) and Penalty-based Boundary Intersection (PBI) scalar optimization functions were tested and for each framework, with only the majority and minority class being considered for objectives. In order to obtain a statistically sound conclusion, the Wilcoxon rank-sum test was run with a significance level of
$\alpha$ = 0.05 on the average hypervolume results. The statistically significant differences between the two frameworks are highlighted in boldface in each of the respective tables.
A row-wise comparison for each of the data sets shows that the semantically ordered approach for Ion, Climate and Glass produced significantly better results when compared to the canonical approach for the hypervolume averages. Yeast$_2$ and Wine had no statistical change in this regard. The results on the Spect and Parkinson's datasets, are mixed and no general conclusions can be drawn. We will explain these results on more detail in the following section. 


At a more granular level, when comparing the semantic ordered approach against its canonical form, the Weighted Sum method performs better in terms of producing more statistically significant improvements, where 5 of the 8 data sets had improved results in regards to their hypervolume averages. The other 3 data sets; Yeast$_2$, Parkinson's and Wine, did not see a statistically significant change. Tchebycheff and Penalty-based Boundary Intersection methods, which make use of the ideal point, produced more mixed results. The Tchebycheff approach performs slightly better in its semantic form where 3 of the 8 data sets (Ion, Climate and Glass) show an improvement versus 1 out of 8 (Spect) in the canonical form however, while the same 3 out of the 8 data sets (Yeast$_2$, Parkinson's and Wine) show an improvement in the semantic approach 4 out of the 8 (Spect, Yeast$_1$, Yeast$_2$ and Parkinson's) are significantly better in their canonical form. The accumulated PO front largely conform with the average hypervolume results, having as good or better performance for the semantic method, although some notable decreases were observed for Climate (TCH and PBI).

\begin{table*}[tbp]
\caption{Average ($\pm$ standard deviation) hypervolume of evolved Pareto-approximated fronts and PO fronts for MOEA/D with WGT, TCH and PBI methods of decomposition for 30 runs.}
\centering
\resizebox{0.90\textwidth}{!}{
\begin{tabular}{ccccccc}\hline
\multirow{3}{*}{Data set} & 
\multicolumn{2}{c}{WGT} &
\multicolumn{2}{c}{TCH} &
\multicolumn{2}{c}{PBI}\\
           & 
           \multicolumn{2}{c}{Hypervolume} & \multicolumn{2}{c}{Hypervolume} & \multicolumn{2}{c}{Hypervolume}\\
    &  Average & PO Front     
    & Average & PO Front 
    & Average & PO Front \\ \hline

Ion &  0.731 $\pm$ 0.030 & 0.854 
    &  0.840 $\pm$ 0.034 & 0.934
    &  0.817 $\pm$ 0.038 & 0.921 \\

Spect  &  0.493 $\pm$ 0.018 & 0.591 
       &  \textbf{0.538} $\pm$ \textbf{0.028} & 0.635
       &  \textbf{0.536} $\pm$ \textbf{0.025} & 0.648 \\ 

Yeast$_1$ &  0.796 $\pm$ 0.015 & 0.852 
          &  0.838 $\pm$ 0.010 & 0.874
          &  \textbf{0.836} $\pm$ \textbf{0.006} & 0.873 \\ 

Yeast$_2$ &  0.887 $\pm$ 0.044 & 0.967 
          &  0.948 $\pm$ 0.008 & 0.975
          &  \textbf{0.939} $\pm$ \textbf{0.024} & 0.976 \\ 
        
Climate &  0.654 $\pm$ 0.054 & 0.863 
     &  0.645 $\pm$ 0.084 & 0.866 
    &  0.603 $\pm$ 0.094 & 0.788 \\

Glass  &  0.758 $\pm$ 0.046 & 0.905 
       &  0.807 $\pm$ 0.051 & 0.925
       &  0.810 $\pm$ 0.048 & 0.917 \\ 

Parkinson's &  0.769 $\pm$ 0.062 & 0.906 
          &  0.788 $\pm$ 0.054 & 0.932
          &  \textbf{0.779} $\pm$ \textbf{0.04}8 & 0.942 \\ 

Wine &  0.915 $\pm$ 0.051 & 0.993
          &  0.959 $\pm$ 0.031 & 1.000
          &  0.954 $\pm$ 0.037 & 0.999 \\ 

\hline
\end{tabular}
}
\label{tab:hypervolumn:moeadcannon}
\end{table*}

\begin{table*}[tbp]
\caption{Average ($\pm$ standard deviation) hypervolume of evolved Pareto-approximated fronts and PO fronts for \textbf{semantically ordered} neighborhood MOEA/D with WGT, TCH and PBI methods of decomposition for 30 runs.}
\centering
\resizebox{0.90\textwidth}{!}{
\begin{tabular}{ccccccc}\hline
\multirow{3}{*}{Data set} & 
\multicolumn{2}{c}{WGT} &
\multicolumn{2}{c}{TCH} &
\multicolumn{2}{c}{PBI}\\
           & 
           \multicolumn{2}{c}{Hypervolume} & \multicolumn{2}{c}{Hypervolume} & \multicolumn{2}{c}{Hypervolume}\\
    &  Average & PO Front     
    & Average & PO Front 
    & Average & PO Front \\ \hline

Ion &  \textbf{0.787} $\pm$ \textbf{0.027} & 0.881 
    &  \textbf{0.854} $\pm$ \textbf{0.018} & 0.927
    &  \textbf{0.849} $\pm$ \textbf{0.020} & 0.937 \\

Spect  &  \textbf{0.509 $\pm$ 0.015} & 0.607 
       &  0.517 $\pm$ 0.021 & 0.605
       &  0.520 $\pm$ 0.021 & 0.615 \\ 

Yeast$_1$ &  \textbf{0.813} $\pm$ \textbf{0.008} & 0.853 
          &  0.832 $\pm$ 0.006 & 0.874
          &  0.834 $\pm$ 0.005 & 0.873 \\ 

Yeast$_2$ &  0.899 $\pm$ 0.025 & 0.961 
          &  0.946 $\pm$ 0.009 & 0.972
          &  0.949 $\pm$ 0.009 & 0.972 \\ 
        
Climate &  \textbf{0.704} $\pm$ \textbf{0.054} & 0.859 
     &  \textbf{0.730} $\pm$ \textbf{0.088} & 0.909 
    &  \textbf{0.763} $\pm$ \textbf{0.068} & 0.908 \\

Glass  &  \textbf{0.793} $\pm$ \textbf{0.036} & 0.907 
       &  \textbf{0.831} $\pm$ \textbf{0.037} & 0.916
       &  \textbf{0.834} $\pm$ \textbf{0.046} & 0.923 \\ 

Parkinson's &  0.771 $\pm$ 0.041 & 0.905
          &  0.802 $\pm$ 0.040 & 0.916
          &  0.814 $\pm$ 0.034 & 0.928 \\ 

Wine &  0.929 $\pm$ 0.027 & 0.998
          &  0.965 $\pm$ 0.021 & 0.997
          &  0.958 $\pm$ 0.023 & 0.998 \\ 

\hline
\end{tabular}
}
\label{tab:hypervolumn:moeadsemantics}
\end{table*}

\subsection{Analysis of objective space}
\label{sec:sub:feasible_solu}

\begin{figure*}
  \centering
  \begin{tabular}{cc}\\
     \multicolumn{2}{c}{Ion}\\
     \hspace{-0.82cm}  
     \includegraphics[width=0.45\textwidth]{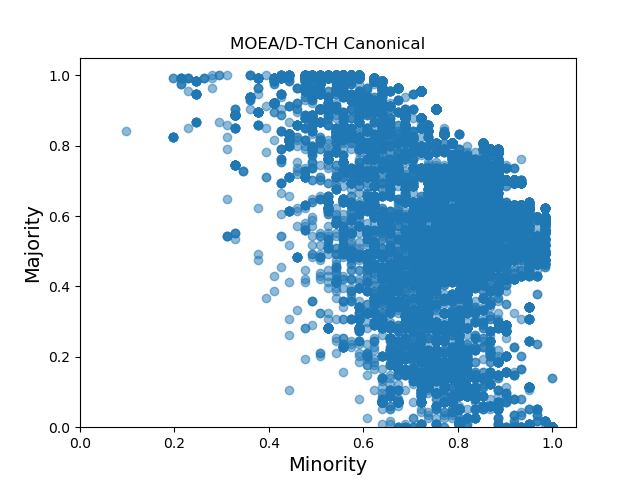}   
     & \hspace{-0.95cm}  
     \includegraphics[width=0.45\textwidth]{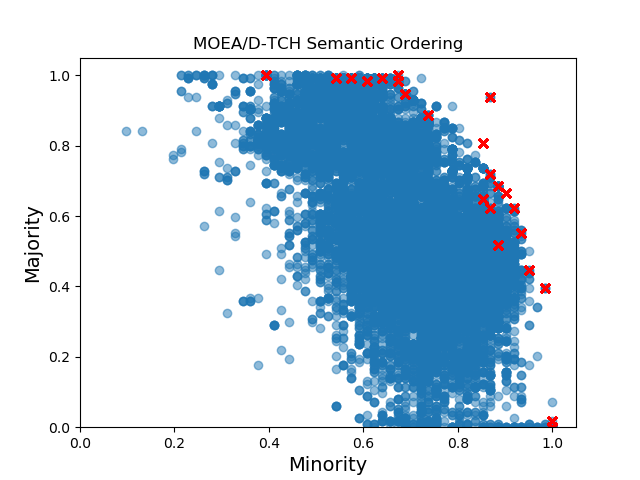}\\
     \multicolumn{2}{c}{Yeast$_1$}\\
     \hspace{-0.82cm}  
     \includegraphics[width=0.45\textwidth]{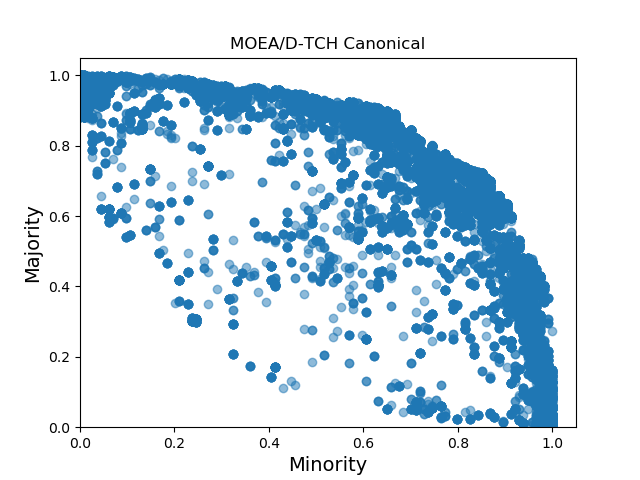}   
     & \hspace{-0.95cm}  
     \includegraphics[width=0.45\textwidth]{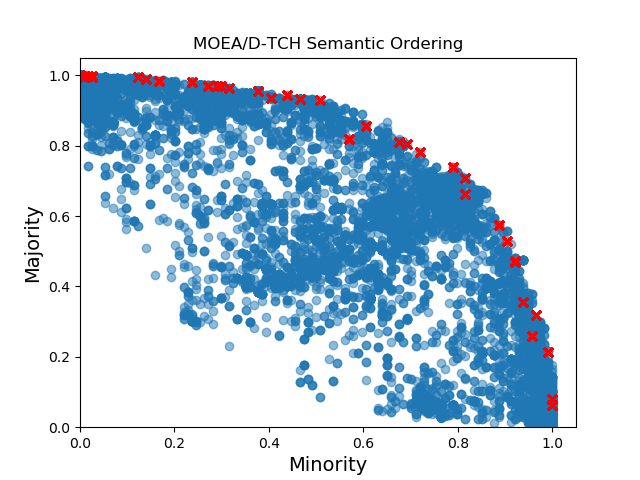} \\

\end{tabular}
\caption{Solutions for every generation for the Ion and Yeast$_1$ data sets. Red cross symbols ('x') denote pivot selections from the external population for all generations for a single run.} 
\label{fig:pivot_per_gen}
\end{figure*}

We have seen some promising results when promoting diversity in MOEA/D algorithms. In other cases, the results are mixed, where canonical MOEA/D methods perform better in a few data sets. To better understand this, we perform a detailed analysis in the objective space. This is shown in  Fig.~\ref{fig:pivot_per_gen}.

The left hand plots show the canonical approach while the right hand plots show the semantic approach. For the semantic approach the selected pivots in the external population for every generation is displayed and denoted with red cross symbols (`x’). All plots are from the same single seeded run and have been selected randomly to avoid having a biased analysis of results. Due to the number of page constraint, we focus our attention on the Ion and Yeast$_1$ data sets, but we see the same trend for the rest of the data sets used in this work.

We can see by comparing the canonical method for the Ion data set (top row) there are a number of discontinuous regions along the border of the feasible space. A notably large region can be observed in the bottom-right hand corner. These areas of sparsity correspond to the pivot selection, with the pivot being selected at either the beginning or end of where these discontinuities occur. When we look at the solutions produced for the same data set but now using the semantic-based method and using the same MOEA/D algorithm (TCH), top right, we can see that discontinuity observed in the canonical method is no longer as prevalent.
In contrast if we look at the Yeast$_1$ data set (bottom row) no such region is observed.

This analysis allows for an intuitive understanding of why the semantic ordering method produced better results for some data sets which make use of an ideal point over others. Data sets which had a relatively smooth boundary along the feasible search space tended to perform better for the canonical method, whereas the data sets that tended to have jagged or discontinuities along the feasible search space tended to perform better for the semantic method. 

\subsection{Analysis of duplication}
\label{sec:sub:duplication}

\begin{figure*}
  \centering
  \
  \begin{tabular}{ccc}
     

          
     \hspace{-0.82cm}  
     \includegraphics[width=0.45\textwidth]{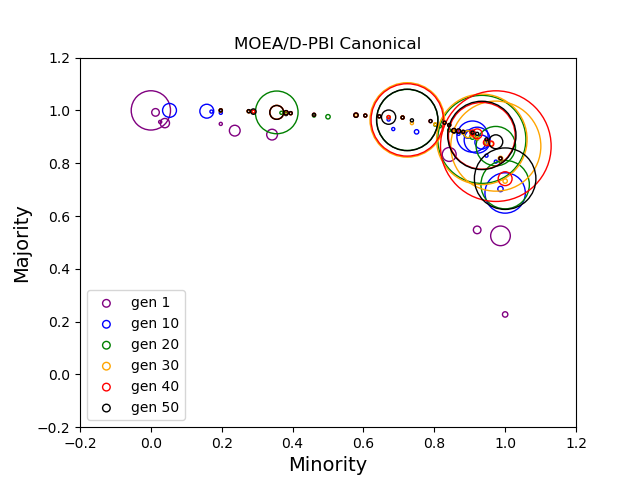}   
     & \hspace{-0.95cm}  
     \includegraphics[width=0.45\textwidth]{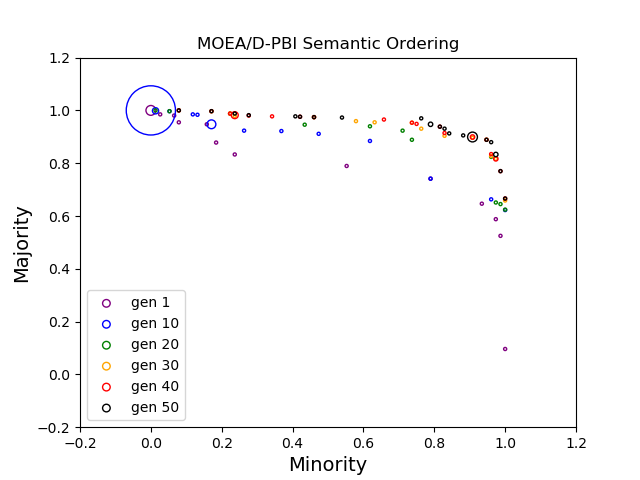} \\

\end{tabular}
\caption{Duplicate frequency of individuals at first Pareto Front for Yeast$_2$ data set with PBI scalar optimization for generations 1, 10, 20, 30, 40 and 50, for a single run.} 
\label{fig:dup_yst2}
\end{figure*}

To demonstrate how the semantic ordering method maintains a lower level of duplication, the external population has been output for Yeast$_2$ data sets at generations 1, 10, 20, 30, 40 and 50 as seen in Fig.~\ref{fig:dup_yst2}. It is important to note that typically these duplicates are removed as part of Step 3.1 as described in Section \ref{sec:sub:moead_canon}. The chosen data set is indicative of the general duplication pattern exhibited by all data sets and to save repetition of results, we have limited our figure to just this data set. The left hand plot and right hand plot show results yielded by the canonical method and the semantic-based method, respectively. The size of each marker represents the number candidate solutions found at that point in the objective space and as such is an indication of the frequency of duplication. For the semantics ordered approach, duplication does not occur as readily but for the canonical approaches their is significant duplication. This detrimental effect is, however, nicely handled by our proposed semantic-based method that encourages diversity.

\section{Conclusion}
\label{sec:conclusions}



This work shows for the first time how semantics can be naturally incorporated into MOEA/D, specifically using the scalar optimization techniques of Weighted Sum, Tchebycheff and PBI methods. The approach referred to as semantic neighborhood ordering reduces the duplication of solutions in objective space, while simultaneously allowing a systematic approach for offspring to compete with neighborhood individuals based on semantics. It was found that for data sets with discontinuous or jagged boundaries along the feasible search space the semantic based method produced significantly better results in terms of the hypervolume averages for the Pareto-approximated fronts. For the boundaries of smooth search spaces, the methods performed as well with only the PBI method producing significantly mixed results. Further studies are necessary to determine how semantics can be of use when using this method.


\section*{Acknowledgments}

\noindent This publication has emanated from research conducted with the financial support of Science Foundation Ireland under Grant number 18/CRT/6049. The authors wish to acknowledge the DJEI/DES/SFI/HEA Irish Centre for High-End Computing (ICHEC) for the provision of computational facilities and support.

\bibliographystyle{abbrv}
\bibliography{semantics}

\begin{thebibliography}{10}

\bibitem{10.1007/978-3-030-41299-9_35}
S.~Azari, B.~Xue, M.~Zhang, and L.~Peng.
\newblock A decomposition based multi-objective genetic programming algorithm
  for classification of highly imbalanced tandem mass spectrometry.
\newblock In S.~Palaiahnakote, G.~Sanniti~di Baja, L.~Wang, and W.~Q. Yan,
  editors, {\em Pattern Recognition}, pages 449--463, Cham, 2020. Springer
  International Publishing.

\bibitem{6198882}
U.~Bhowan, M.~Johnston, M.~Zhang, and X.~Yao.
\newblock Evolving diverse ensembles using genetic programming for
  classification with unbalanced data.
\newblock {\em IEEE Transactions on Evolutionary Computation}, 17(3):368--386,
  June 2013.

\bibitem{Deb:2001:MOU:559152}
K.~Deb.
\newblock {\em Multi-Objective Optimization Using Evolutionary Algorithms}.
\newblock John Wiley \& Sons, Inc., New York, NY, USA, 2001.

\bibitem{Dua:2019}
D.~Dua and C.~Graff.
\newblock {UCI} machine learning repository, 2017.

\bibitem{galvan2021neuroevolution}
{E. Galván and P. Mooney}.
\newblock {Neuroevolution in Deep Neural Networks: Current Trends and Future
  Challenges}.
\newblock arXiv preprint arXiv: 2006.05415, 2020.

\bibitem{Eiben:2015:nature}
A.~E. Eiben and J.~Smith.
\newblock From evolutionary computation to the evolution of things.
\newblock {\em Nature}, 521:476--482, 28 May 2015.

\bibitem{EibenBook2003}
A.~E. Eiben and J.~E. Smith.
\newblock {\em Introduction to {E}volutionary {C}omputing}.
\newblock Springer Verlag, 2003.

\bibitem{DBLP:conf/gecco/GalvanS19}
E.~Galv{\'{a}}n and M.~Schoenauer.
\newblock Promoting semantic diversity in multi-objective genetic programming.
\newblock In A.~Auger and T.~St{\"{u}}tzle, editors, {\em Proceedings of the
  Genetic and Evolutionary Computation Conference, {GECCO} 2019, Prague, Czech
  Republic, July 13-17, 2019}, pages 1021--1029. {ACM}, 2019.

\bibitem{galvan2020promoting}
E.~Galv{\'a}n and F.~Stapleton.
\newblock Promoting semantics in multi-objective genetic programming based on
  decomposition.
\newblock {\em arXiv preprint arXiv:2012.04717}, 2020.

\bibitem{9308386}
E.~Galv{\'a}n and F.~Stapleton.
\newblock Semantic-based distance approaches in multi-objective genetic
  programming.
\newblock In {\em 2020 IEEE Symposium Series on Computational Intelligence
  (SSCI)}, pages 149--156. IEEE, 2020.

\bibitem{DBLP:conf/cec/LopezCTK13}
E.~Galv{\'{a}}n-L{\'{o}}pez, B.~Cody{-}Kenny, L.~Trujillo, and A.~Kattan.
\newblock Using semantics in the selection mechanism in genetic programming:
  {A} simple method for promoting semantic diversity.
\newblock In {\em Proceedings of the {IEEE} Congress on Evolutionary
  Computation, {CEC} 2013, Cancun, Mexico, June 20-23, 2013}, pages 2972--2979,
  2013.

\bibitem{galvan_neurocomputing_2015}
E.~Galv{\'a}n-L{\'o}pez, T.~Curran, J.~McDermott, and P.~Carroll.
\newblock Design of an autonomous intelligent demand-side management system
  using stochastic optimisation evolutionary algorithms.
\newblock {\em Neurocomputing}, 170:270--285, 2015.

\bibitem{DBLP:conf/eurogp/LopezDP08}
E.~Galv{\'{a}}n-L{\'{o}}pez, S.~Dignum, and R.~Poli.
\newblock The effects of constant neutrality on performance and problem
  hardness in {GP}.
\newblock In M.~O'Neill, L.~Vanneschi, S.~M. Gustafson,
  A.~Esparcia{-}Alc{\'{a}}zar, I.~D. Falco, A.~D. Cioppa, and E.~Tarantino,
  editors, {\em Genetic Programming, 11th European Conference, EuroGP 2008,
  Naples, Italy, March 26-28, 2008. Proceedings}, volume 4971 of {\em Lecture
  Notes in Computer Science}, pages 312--324. Springer, 2008.

\bibitem{DBLP:conf/cec/LopezMOB10}
E.~Galv{\'a}n-L{\'o}pez, J.~McDermott, M.~O'Neill, and A.~Brabazon.
\newblock Defining locality in genetic programming to predict performance.
\newblock In {\em Proceedings of the {IEEE} Congress on Evolutionary
  Computation, {CEC} 2010, Barcelona, Spain, 18-23 July 2010}, pages 1--8,
  2010.

\bibitem{Galvan-Lopez:2010:TUL:1830483.1830646}
E.~Galv\'{a}n-L\'{o}pez, J.~McDermott, M.~O'Neill, and A.~Brabazon.
\newblock Towards an understanding of locality in genetic programming.
\newblock In {\em Proceedings of the 12th Annual Conference on Genetic and
  Evolutionary Computation}, GECCO '10, pages 901--908, New York, NY, USA,
  2010. ACM.

\bibitem{Galvan-Lopez2011}
E.~Galv{\'a}n-L{\'o}pez, J.~McDermott, M.~O'Neill, and A.~Brabazon.
\newblock Defining locality as a problem difficulty measure in genetic
  programming.
\newblock {\em Genetic Programming and Evolvable Machines}, 12(4):365--401,
  2011.

\bibitem{Galvan-Lopez2016}
E.~Galv{\'a}n-L{\'o}pez, E.~Mezura-Montes, O.~Ait~ElHara, and M.~Schoenauer.
\newblock On the use of semantics in multi-objective genetic programming.
\newblock In J.~Handl et~al., editors, {\em Parallel Problem Solving from
  Nature -- PPSN XIV: 14th International Conference, Edinburgh, UK, September
  17-21, 2016, Proceedings}, pages 353--363. Springer, 2016.

\bibitem{DBLP:conf/ppsn/LopezP06_2}
E.~Galv{\'{a}n}-L{\'{o}}pez and R.~Poli.
\newblock Some steps towards understanding how neutrality affects evolutionary
  search.
\newblock In T.~P. Runarsson, H.~Beyer, E.~K. Burke, J.~J. {Merelo
  Guerv{'{o}}s}, L.~D. Whitley, and X.~Yao, editors, {\em Parallel Problem
  Solving from Nature - {PPSN} IX, 9th International Conference, Reykjavik,
  Iceland, September 9-13, 2006, Procedings}, volume 4193, pages 778--787.
  Springer, 2006.

\bibitem{DBLP:conf/eurogp/LopezPC04}
E.~Galv{\'{a}}n-L{\'{o}}pez, R.~Poli, and C.~A.~C. Coello.
\newblock Reusing code in genetic programming.
\newblock In M.~Keijzer, U.~O'Reilly, S.~M. Lucas, E.~Costa, and T.~Soule,
  editors, {\em Genetic Programming, 7th European Conference, EuroGP2004,
  Coimbra, Portugal, April 5-7, 2004, Proceedings}, volume 3003 of {\em Lecture
  Notes in Computer Science}, pages 359--368. Springer, 2004.

\bibitem{DBLP:journals/evs/LopezPKOB11}
E.~Galv{\'{a}}n-L{\'{o}}pez, R.~Poli, A.~Kattan, M.~O'Neill, and A.~Brabazon.
\newblock Neutrality in evolutionary algorithms... what do we know?
\newblock {\em Evolving Systems}, 2(3):145--163, 2011.

\bibitem{Galvan_SAC_2014}
E.~Galv\'an-L\'opez, A.~Taylor, S.~Clarke, and V.~Cahill.
\newblock Design of an automatic demand-side management system based on
  evolutionary algorithms.
\newblock In {\em Proceedings of the 29th Annual Symposium on Applied
  Computing, SAC '14}, pages 525 -- 530, Gyeongju, Korea, 26-28 Mar. 2014. ACM.

\bibitem{Galvan_MICAI_2016}
E.~Galv\'an-L\'opez, L.~V\'azquez-Mendoza, and L.~Trujillo.
\newblock Stochastic semantic-based multi-objective genetic programming
  optimisation for classification of imbalanced data.
\newblock In O.~Pichardo-Lagunas and S.~Miranda-Jim\'enez, editors, {\em
  Advances in Soft Computing}, chapter~22, pages 261--272. Springer, 2016.

\bibitem{galvan2021neuroevolutionb}
E.~Galván.
\newblock Neuroevolution in deep learning: The role of neutrality.
\newblock arXiv preprint arXiv: 2102.08475, 2021.

\bibitem{Deb02afast}
S.~A. K.~Deb, A.~Pratap and T.~Meyarivan.
\newblock A fast and elitist multiobjective genetic algorithm: Nsga-ii.
\newblock {\em IEEE Transactions on Evolutionary Computation}, 6:182--197,
  2002.

\bibitem{Koza:1992:GPP:138936}
J.~R. Koza.
\newblock {\em Genetic Programming: On the Programming of Computers by Means of
  Natural Selection}.
\newblock MIT Press, Cambridge, MA, USA, 1992.

\bibitem{Little_2007}
M.~A. Little, P.~E. McSharry, S.~J. Roberts, D.~A. Costello, and I.~M. Moroz.
\newblock Exploiting nonlinear recurrence and fractal scaling properties for
  voice disorder detection.
\newblock {\em BioMedical Engineering OnLine}, 6(1):23, 2007.

\bibitem{gmd-6-1157-2013}
D.~D. Lucas, R.~Klein, J.~Tannahill, D.~Ivanova, S.~Brandon, D.~Domyancic, and
  Y.~Zhang.
\newblock Failure analysis of parameter-induced simulation crashes in climate
  models.
\newblock {\em Geoscientific Model Development}, 6(4):1157--1171, 2013.

\bibitem{McPhee:2008:SBB:1792694.1792707}
N.~F. McPhee, B.~Ohs, and T.~Hutchison.
\newblock Semantic building blocks in genetic programming.
\newblock In {\em Proceedings of the 11th European conference on Genetic
  programming}, EuroGP'08, pages 134--145, Berlin, Heidelberg, 2008.
  Springer-Verlag.

\bibitem{DBLP:conf/ppsn/MoraglioKJ12}
A.~Moraglio, K.~Krawiec, and C.~G. Johnson.
\newblock Geometric semantic genetic programming.
\newblock In C.~A.~C. Coello, V.~Cutello, K.~Deb, S.~Forrest, G.~Nicosia, and
  M.~Pavone, editors, {\em PPSN (1)}, volume 7491 of {\em LNCS}, pages 21--31.
  Springer, 2012.

\bibitem{6808504}
T.~P. Pawlak, B.~Wieloch, and K.~Krawiec.
\newblock Semantic backpropagation for designing search operators in genetic
  programming.
\newblock {\em IEEE Transactions on Evolutionary Computation}, 19(3):326--340,
  June 2015.

\bibitem{10.1007/978-3-540-73482-6_9}
R.~Poli and E.~Galv{\'a}n-L{\'o}pez.
\newblock On the effects of bit-wise neutrality on fitness distance
  correlation, phenotypic mutation rates and problem hardness.
\newblock In C.~R. Stephens, M.~Toussaint, D.~Whitley, and P.~F. Stadler,
  editors, {\em Foundations of Genetic Algorithms}, pages 138--164, Berlin,
  Heidelberg, 2007. Springer Berlin Heidelberg.

\bibitem{DBLP:journals/tec/PoliL12}
R.~Poli and E.~Galv{\'{a}}n-L{\'{o}}pez.
\newblock The effects of constant and bit-wise neutrality on problem hardness,
  fitness distance correlation and phenotypic mutation rates.
\newblock {\em {IEEE} Trans. Evolutionary Computation}, 16(2):279--300, 2012.

\bibitem{10.1145/3377929.3397486}
S.~Ruberto, V.~Terragni, and J.~H. Moore.
\newblock Sgp-dt: Towards effective symbolic regression with a semantic gp
  approach based on dynamic targets.
\newblock In {\em Proceedings of the 2020 Genetic and Evolutionary Computation
  Conference Companion}, GECCO '20, page 25–26, New York, NY, USA, 2020.
  Association for Computing Machinery.

\bibitem{Sato2016chain}
H.~Sato.
\newblock Chain-reaction solution update in moea/d and its effects on multi-
  and many-objective optimization.
\newblock {\em Soft Computing}, 20, 10 2016.

\bibitem{stapleton2021msc}
F.~Stapleton.
\newblock Improving semantic diversity in multi-objective genetic programming.
\newblock MSc Thesis. Maynooth University, 2020.

\bibitem{10.1007/978-3-642-04962-0_7}
N.~Q. Uy, N.~X. Hoai, M.~O'Neill, B.~McKay, and E.~Galv{\'a}n-L{\'o}pez.
\newblock An analysis of semantic aware crossover.
\newblock In Z.~Cai, Z.~Li, Z.~Kang, and Y.~Liu, editors, {\em Computational
  Intelligence and Intelligent Systems}, pages 56--65, Berlin, Heidelberg,
  2009. Springer Berlin Heidelberg.

\bibitem{Uy2011}
N.~Q. Uy, N.~X. Hoai, M.~O'Neill, R.~I. McKay, and E.~Galv{\'a}n-L{\'o}pez.
\newblock Semantically-based crossover in genetic programming: application to
  real-valued symbolic regression.
\newblock {\em Genetic Programming and Evolvable Machines}, 12(2):91--119,
  2011.

\bibitem{DBLP:conf/ae/UyOHML09}
N.~Q. Uy, M.~O'Neill, N.~X. Hoai, B.~McKay, and E.~Galv{\'{a}}n-L{\'{o}}pez.
\newblock Semantic similarity based crossover in {GP:} the case for real-valued
  function regression.
\newblock In P.~Collet, N.~Monmarch{\'{e}}, P.~Legrand, M.~Schoenauer, and
  E.~Lutton, editors, {\em Artifical Evolution, 9th International Conference,
  Evolution Artificielle, EA, 2009, Strasbourg, France, October 26-28, 2009.
  Revised Selected Papers}, volume 5975 of {\em Lecture Notes in Computer
  Science}, pages 170--181. Springer, 2009.

\bibitem{Vanneschi2013}
L.~Vanneschi, M.~Castelli, L.~Manzoni, and S.~Silva.
\newblock {\em A New Implementation of Geometric Semantic GP and Its
  Application to Problems in Pharmacokinetics}, pages 205--216.
\newblock Springer Berlin Heidelberg, Berlin, Heidelberg, 2013.

\bibitem{Vanneschi:2014:SSM:2618052.2618082}
L.~Vanneschi, M.~Castelli, and S.~Silva.
\newblock A survey of semantic methods in genetic programming.
\newblock {\em Genetic Programming and Evolvable Machines}, 15(2):195--214,
  June 2014.

\bibitem{4358754}
Q.~{Zhang} and H.~{Li}.
\newblock Moea/d: A multiobjective evolutionary algorithm based on
  decomposition.
\newblock {\em IEEE Transactions on Evolutionary Computation}, 11(6):712--731,
  2007.

\bibitem{Zitzler01spea2:improving}
E.~Zitzler, M.~Laumanns, and L.~Thiele.
\newblock Spea2: Improving the strength pareto evolutionary algorithm.
\newblock Technical report, Swiss Federal Institute of Technology Zurich (ETH),
  2001.

\end{thebibliography}

\end{document}